# A Novel Approach to Multimedia Ontology Engineering for Automated Reasoning over Audiovisual LOD Datasets

Leslie F. Sikos

Flinders University, Adelaide, Australia

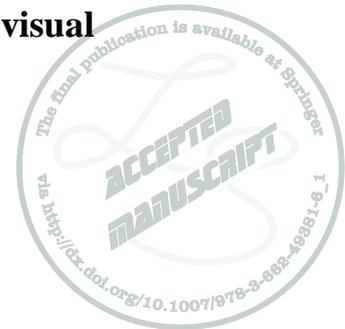

**Abstract.**
Multimedia reasoning, which is suitable for, among others, multimedia content analysis and high-level video scene interpretation, relies on the formal and comprehensive conceptualization of the represented knowledge domain. However, most multimedia ontologies are not exhaustive in terms of role definitions, and do not incorporate complex role inclusions and role interdependencies. In fact, most multimedia ontologies do not have a role box at all, and implement only a basic subset of the available logical constructors. Consequently, their application in multimedia reasoning is limited. To address the above issues, VidOnt, the very first multimedia ontology with $\mathcal{SROIQ}^{(D)}$ expressivity and a DL-safe ruleset has been introduced for next-generation multimedia reasoning. In contrast to the common practice, the formal grounding has been set in one of the most expressive description logics, and the ontology validated with industry-leading reasoners, namely HermiT and FaCT++. This paper also presents best practices for developing multimedia ontologies, based on my ontology engineering approach.

## 1 Introduction to Multimedia Reasoning

*Description logics* (DL), which are formal knowledge representation languages, have early implementations in the multimedia domain since the 1990s [1]. They are suitable for the expressive formalization of multimedia contents and the semantic refinement of video segmentation [2]. DL-based knowledge representations, such as OWL ontologies, can serve as the basis for multimedia content analysis [3], event detection [4], high-level video scene interpretation [5], abductive reasoning to differentiate between similar concepts in image sequence interpretation [6], and constructing high-level media descriptors [7], particularly if the ontology contains not only terminological and assertional axioms (that form a knowledge base), but also a role box and a ruleset. Ontology rules make video content understanding possible and improve the quality of structured annotations of concepts and predicates [8]. Natural language processing algorithms can be used to curate the represented video concepts while preserving provenance data, and assist to achieve consistency in multimedia ontologies [9].

In contrast to ontologies of other knowledge domains, video ontologies need a specific set of motion events to represent spatial changes of video scenes, which are characterized by subconcepts, multiple interpretations, and ambiguity [10]. Research results in structured video annotations are particularly promising for constrained videos, where the knowledge domain is known, such as medical videos, news videos, tennis videos, and soccer videos [11].

In spite of the benefits of multimedia reasoning in video scene interpretation and understanding, most multimedia ontologies lack the expressivity and constructors necessary for complex inference tasks [12]. To address the reasoning limitations of multimedia ontologies, the VidOnt ontology has been introduced, which exploits all mathematical constructors of the underlying expressive description logic, and features a role box and a ruleset missing from previous multimedia ontologies for automated scene interpretation and video understanding [13]. VidOnt is suitable for the knowledge representation and lightweight annotation of objects and actors depicted in videos, providing technical, licensing, and general metadata as structured data, as well as for multimedia reasoning and Linked Open Data (LOD) interlinking.

## 2 Formalism with Description Logics

The majority of web ontologies written in the Web Ontology Language (OWL) are implementations of a description logic [14]. Description logics are decidable fragments of first-order logic (FOL): DL concepts are equivalent to FOL unary predicates, DL roles to FOL binary predicates, DL individuals to FOL constants, DL concept expressions to FOL formulae with one free variable, role expressions to FOL formulae with two free variables, and so on. Description logics are more efficient in decision problems than first-order predicate logic (which uses predicates and quantified variables over non-logical objects) and more expressive than propositional logic (which uses declarative propositions and does not use quantifiers). A description logic can efficiently model concepts, roles, individuals, and their relationships.

**Definition 1 (Concept).** The concept $C$ of an ontology is defined as a pair that can be expressed as $C = (X^C, Y^C)$, wherein $X^C \subseteq X$ is a set of attributes describing the concept, and $Y^C \subseteq Y$ is the domain of the attributes, $Y^C = \bigcup_{x \in X^C} Y_x$

**Definition 2 (Role).** A role is either $r \in N_R$, an inverse role $r^-$ with $r \in N_R$, or a universal role $U^1$.

A core modeling concept of a description logic is the *axiom*, which is a logical statement about the relation between roles and/or concepts.

**Definition 3 (Axiom).** An axiom is either
- a general concept inclusion of the form $A \sqsubseteq B$ for concepts $A$ and $B$, or
- an individual assertion of one of the forms $a : C$, $(a, b) : R$, $(a, b) : \neg r$, $a = b$ or $a \neq b$ for individuals $a$, $b$ and a role $r$, or
- a role assertion of one of the forms $R \sqsubseteq S$, $R_1 \circ \ldots \circ R_n \sqsubseteq S$, $\text{Asy}(R)$, $\text{Ref}(R)$, $\text{Irr}(R)$, $\text{Dis}(R, S)$ for roles $R$, $R_i$, $S$.

After determining the domain and scope of the ontology, and potential term reuse from external ontologies, the terms of the knowledge domain are enumerated, followed by the creation of the class hierarchy, the concept and predicate definitions, their relationships, and individuals. Both the first-order logic and the description logic syntax correspond to OWL, so axioms written in either syntax can be translated to the desired OWL serialization, such as Turtle, as demonstrated in Table 1.

**Table 1.** Description Logic to OWL 2 DL Translation Examples

| DL Axiom | Turtle Syntax |
| --- | --- |
| $a \approx b$ | `a owl:sameAs b .` |
| $a \not\approx b$ | `a owl:differentFrom b .` |
| $C \sqsubseteq D$ | `C rdfs:subClassOf D .` |
| $C(a)$ | `a rdf:type C .` |
| $r(a, b)$ | `a r b .` |
| $r^-(a, b)$ | `b r a .` |

The data model of the VidOnt ontology has been formalized in the very expressive yet decidable $\mathcal{SROIQ}^{(D)}$ description logic, which exploits all constructors of OWL 2 DL from concept constructors to complex role inclusion axioms, as will be discussed in the following sections.

**Definition 4 ($\mathcal{SROIQ}$ ontology).** A $\mathcal{SROIQ}$ ontology is a set $\mathcal{O}$ of axioms including $\varrho \sqsubseteq R$ complex role inclusions, $\text{Dis}(S_1, S_2)$ disjoint roles, $C \sqsubseteq D$ concept inclusions, $C(a)$ concept assertions, and $R(a, b)$ role assertions, wherein $\varrho$ is a role chain, $R_{(i)}$ and $S_{(i)}$ are roles, $C$ and $D$ are concepts, and $a$, $b$ individuals, such that the set of all role inclusion axioms in $\mathcal{O}$ are $\prec$-regular for some regular order $\prec$ on roles.

---

[1] In the $\mathcal{SROIQ}$ description logic. Many less expressive DLs do not provide inverse roles, and no other ontology supports the universal role, which has been introduced in $\mathcal{SROIQ}$.

### 2.1 Concept Constructors

The $\mathcal{SROIQ}$ description logic supports a wide range of concept expression constructors, including concept assertion, conjunction, disjunction, complement, top concept, bottom concept, role restrictions (existential and universal restrictions), number restrictions (at-least and at-most restrictions), local reflexivity, and nominals.

**Definition 5 ($\mathcal{SROIQ}$ concept expression).** A set of $\mathcal{SROIQ}$ concept expressions is defined as $\mathbf{C} ::= N_C \mid (C \sqcap C) \mid (C \sqcup C) \mid \neg C \mid \top \mid \bot \mid \exists R.C \mid \forall R.C \mid \geqslant nR.C \mid \leqslant nR.C \mid \exists R.Self \mid \{N_I\}$, wherein C represents concepts, R is a set of roles, and $n$ is a non-negative integer.

### 2.2 Axioms

VidOnt defines terminological, assertional, and relational axioms. As you will see, constructors not exploited in previously released multimedia ontologies, in particular the role box axioms, significantly extend the application potential in data integration, knowledge management, and multimedia reasoning.

#### 2.2.1 TBox Axioms

The concepts and roles of VidOnt have been defined in a hierarchy incorporating de facto standard structured definitions, and can be deployed in fully-featured knowledge representations in an RDF serialization, such as Turtle or RDF/XML, or as lightweight markup annotations in HTML5 Microdata, JSON-LD, or RDFa. Terminological knowledge is included in VidOnt by defining the relationship of classes and properties as subclass axioms and subproperty axioms, respectively, and specifying domains and ranges for the properties. The TBox axioms leverage constructors such as subclass relationships ($\sqsubseteq$), equivalence ($\equiv$), conjunction ($\sqcap$), and disjunction ($\sqcup$), negation ($\neg$), property restrictions ($\forall$, $\exists$), tautology ($\top$), and contradiction ($\bot$).

**Definition 6 (TBox).** A TBox $\mathcal{T}$ is a finite collection of concept inclusion axioms in the form $C \sqsubseteq D$ and concept equivalence axioms in the form $C \equiv D$, wherein $C$ and $D$ are concepts.

For example, TBox axioms can express that live action is a movie type, or narrators are equivalent to lectors, as shown in Table 2.

**Table 2.** Expressing Terminological Knowledge with TBox Axioms

| DL Syntax | Turtle Syntax |
|---|---|
| liveAction $\sqsubseteq$ Movie | `:liveAction rdfs:subClassOf :Movie .` |
| remakeOf $\sqsubseteq$ basedOn | `:remakeOf rdfs:subPropertyOf :basedOn .` |
| Narrator $\equiv$ Lector | `:Narrator owl:equivalentClass :Lector .` |

### 2.2.2 ABox Axioms

Individuals and their relationships are represented using ABox axioms.

**Definition 7 (ABox).** An ABox $\mathcal{A}$ is a finite collection of axioms of the form x:D, ⟨x, y⟩:R, where x and y are individual names, D is a concept, and R is a role. An individual assertion can be
- a concept assertion, $C(a)$
- a role assertion, $r(a, b)$, or a negated role assertion, $\neg r(a, b)$
- an equality statement, $a \approx b$
- an inequality statement, $a \not\approx b$

wherein a, b ∈ $N_I$ individual names, $C \in C$ a concept expression, and $r \in R$ a role, each of which is demonstrated in Table 3.

**Table 3.** Asserting Individuals with ABox Axioms

| DL Syntax | Turtle Syntax |
|---|---|
| computerAnimation(Zambezia) | `:Zambezia a :computerAnimation .` |
| directedBy(Unforgiven, ClintEastwood) | `:Unforgiven :directedBy :ClintEastwood .` |
| 房仕龍 ≈ JackieChan | `:房仕龍 owl:sameIndividualAs :JackieChan .` |
| RobinWilliams $\not\approx$ RobbieWilliams | `:RobinWilliams owl:differentFrom :RobbieWilliams .` |

### 2.2.3 RBox Axioms

Most multimedia ontologies define terminological and assertional axioms only, which form a knowledge base only, rather than a fully-featured ontology.

**Definition 8 (Knowledge Base).** A DL knowledge base $\mathcal{K}$ is a pair ⟨$\mathcal{T}$, $\mathcal{A}$⟩ where
- $\mathcal{T}$ is a set of terminological axioms (TBox)
- $\mathcal{A}$ is a set of assertional axioms (ABox)

Beyond Abox and TBox axioms, $\mathcal{SROIQ}$ also supports *role box* (*RBox*) axioms to collect all statements related to roles and the interdependencies between roles, which is particularly useful for multimedia reasoning.

**Definition 9 (RBox).** A role box (RBox) $\mathcal{R}$ is a role hierarchy, a finite collection of generalized role inclusion axioms of the form $R \sqsubseteq S$, role equivalence axioms in the form $R \equiv S$, complex role inclusions in the form $R_1 \circ R_2 \sqsubseteq S$, and role disjointness declarations in the form Dis($R$, $S$), wherein $R$ and $S$ are roles, and transitivity axioms of the form $R^+ \sqsubseteq R$, wherein $R^+$ is a set of transitive roles.

Some examples for role box axioms are shown in Table 4.

Table 4. Modeling Relationships between Roles with RBox Axioms

| DL Syntax | Turtle Syntax |
|---|---|
| starredIn ∘ starredIn ⊑ co-starred | `:co-starred owl:propertyChainAxiom (:starredIn :starredIn) .` |
| Dis(parentOf, childOf) | `:x a owl:AllDisjointProperties ; owl:members (:parentOf :childOf) .` |
| basedOn ∘ basedOn ⊑ basedOn | `:basedOn a owl:TransitiveProperty .` |

### 2.3 DL-Safe Ruleset

While $\mathcal{SROIQ}^{(\mathcal{D})}$, the description logic of OWL 2 DL, is very expressive, it can only express axioms of a certain tree structure, because OWL 2 DL corresponds to a decidable subset of first-order predicate logic. There are decidable rule-based formalisms, such as function-free Horn rules, which are not restricted in this regard.

**Definition 10 (Rule).** A *rule* R is given as $H \leftarrow B_1, \ldots, B_n (n \geq 0)$, wherein $H, B_1, \ldots, B_n$ are atoms, $H$ is called the *head* (conclusion or consequent) and $B_1, \ldots, B_n$ the *body* (premise or antecedent).

While some OWL 2 axioms correspond to rules, such as class inclusion and property inclusion, some classes can be decomposed as rules, and property chain axioms provide rule-like axioms, there are rules that cannot be expressed in OWL 2 rules. For example, a rule head with two variables cannot be represented as a subclass axiom, or a rule body that contains a class expression cannot be described by a subproperty axiom. To add the additional expressivity of rules to OWL 2 DL, ontologies can be extended with *SWRL*[2] rules which, however, make ontologies undecidable. The solution is to apply *DL-safe rules*, wherein each variable must occur in a non-DL-atom in the rule body [15], i.e., DL-safe rules can be considered SWRL rules restricted to known individuals. DL-safe rules are very expressive and decidable at the same time.

**Definition 11 (DL-safe rule).** Let *KB* be a $\mathcal{SROIQ}^{(\mathcal{D})}$ knowledge base, and let $N_P$ be a set of predicate symbols such that $N_C \cup N_{R_a} \cup N_{R_c} \subseteq N_P$. A DL-atom is an atom of the form $A(s)$, where $A \in N_C$, or of the form $r(s, t)$, where $r \in N_{R_a} \cup N_{R_c}$. A rule $r$ is called DL-safe if each variable in $r$ occurs in a non-DL-atom in the rule body.

As an example, assume we have axioms to define award-winning actors (1–4).

$$\text{AwardWinnerActor} \equiv \text{won}.\exists\text{Award} \quad (1)$$

$$\text{Actor}(a), \text{Actor}(b), \text{Actor}(c) \quad (2)$$

$$\text{Award}(d) \quad (3)$$

$$\text{won}(a, d) \quad (4)$$

---

[2] Semantic Web Rule Language

Based on the axioms, a DL-safe rule can be written to infer new assertional axioms (5).

$$\text{AwardWinnerActor}(x) \leftarrow \text{won}(?x, ?y) \quad (5)$$

Using the above rule (5), reasoners can infer that actor *a* is an award winner (6).

$$\text{AwardWinnerActor}(a) \quad (6)$$

Without a DL-safe restriction containing special non-DL literals O(*x*) and O(*y*) in the rule body and the assertion of each individual, reasoners would assert that actors *a*, *b*, and *c* are award winners (7).

$$\text{AwardWinnerActor}(a), \text{AwardWinnerActor}(b), \text{AwardWinnerActor}(c) \quad (7)$$

## 3 Multimedia Reasoning

The feasibility and efficiency of automated reasoning relies on the accurate conceptualization and comprehensive description of relations between concepts, predicates, and individuals [16]. Advanced reasoning is infeasible without expressive constructors, most of which are not implemented in multimedia ontologies other than VidOnt. For example, the Visual Descriptor Ontology (VDO), which was published as an "ontology for multimedia reasoning" [17], has in fact very limited description logic expressivity (corresponding to $\mathcal{ALH}$) and reasoning potential. In the next sections we compare TBox and ABox reasoning supported by most ontologies to Rbox and rule-based reasoning not supported by any multimedia ontology except VidOnt.

### 3.1 Tableau-Based Consistency Checking

Most OWL-reasoners, such as FaCT++, Pellet, and RacerPro, are based on *tableau* algorithms. They attempt to construct a model that satisfies all axioms of an ontology to prove (un)satisfiability. Based on the ABox axioms, a set of elements is created, which is used to retrieve concept memberships and role assertions. Typically, the constructed intermediate model does not satisfy all TBox and RBox axioms, so the model is updated accordingly with each iteration. As a result, new concept memberships and role relationships might be generated. When a case distinction occurs, the algorithm might have to backtrack. If a state is reached where all axioms are satisfied, the ontology is considered satisfiable. OWL 2 reasoners, such as HermiT, usually use a tableau refinement based on the *hypertableau* and *hyperresolution* calculi to reduce the nondeterminism caused by general inclusion axioms [18].

To demonstrate integrity checking with reasoning, assume the following axioms:

$$\text{acts} \sqsubseteq \text{lives} \quad (8)$$

$$\text{canAct} \sqsubseteq \neg\text{DeadActor} \quad (9)$$

$$\text{Actor} \sqsubseteq \text{DeadActor} \sqcup \text{LivingActor} \quad (10)$$

$$\text{activeActor} \sqsubseteq \text{lives.Actor} \sqcap \forall \text{lives.canAct} \quad (11)$$

$$\text{activeActor}(a) \quad (12)$$

Based on the only ABox axiom (12), tableau-based reasoners would assume that *a* is an active actor, which would not satisfy the definition of living actors (11). Next, reasoners would introduce a new concept which logically corresponds to the Person concept. The connection between the individual (*a*) and the new concept (Person) is defined with the acts predicate. As a result, the definition of active actors (11) is now satisfied, however, other TBox axioms are invalidated (8 and 10). To address this issue, reasoners would introduce a lives connection between individual *a* and the Person concept. Finally, a case distinction is needed, because a person can be either dead (DeadActor) or alive (LivingActor). In the first case, (11) is violated because of the second part of its consequence. To address this issue, Person has to be marked with canAct, which in turn invalidates (9), meaning that Person must be ¬DeadActor. Because Person cannot be marked with both DeadActor and ¬DeadActor, the algorithm needs to backtrack. In the second case, Person is marked as LivingActor, which violates (11), so Person must be marked with canAct, which invalidates (9). Consequently, Person is marked as ¬DeadActor, which leads to a state with a knowledge representation model satisfying all axioms, upon which reasoners can conclude that the ontology is satisfiable.

### 3.2  RBox and Rule-Based Reasoning over Audiovisual Contents

Take a simplistic example which combines RBox reasoning with rule-based reasoning not supported by any other multimedia ontology but VidOnt, to infer statements that are not explicitly defined. Assume the following base ontology:

$$\text{Actor}(a), \text{Actor}(b), \text{Actor}(c), \text{Actor}(d) \quad (13)$$

$$\text{Movie}(m), \text{Series}(s), \text{partOf}(m, s) \quad (14)$$

$$\text{partOf} \circ \text{starredIn} \sqsubseteq \text{co-starredWith} \quad (15)$$

$$\text{starredIn}(a, m), \text{starredIn}(b, m), \text{starredIn}(c, m), \text{starredIn}(d, s) \quad (16)$$

Also assume the following rule:

$$\text{starredIn}(?x, m) \rightarrow \text{co-starredWith}(?x, d) \quad (17)$$

Based on the ABox and TBox axioms (13, 14, 16) and the DL-safe rule (17), reasoners can generate new object property assertions about the actors who co-starred with actor *d* (24–26):

$$\text{co-starredWith}(a, d), \text{co-starredWith}(b, d), \text{co-starredWith}(c, d) \quad (18)$$

Furthermore, based on the property chain axiom (15), it can be inferred that actors who starred in at least one part of a series appeared in the series (19):

$$\text{starredIn}(a, s), \text{starredIn}(b, s), \text{starredIn}(c, s) \quad (19)$$

The resulting axioms are automatically generated with full certainty, making the combination of complex role inclusion axioms and DL-safe rules suitable for big data implementations where manual annotation is not an option, for video cataloging to automatically generate new axioms through user or programmatic queries, and for knowledge discovery, such as identifying factors from medical videos that, when occur together, indicate a serious condition or disease.

## Conclusions and Future Work

Multimedia ontology engineers often apply a bottom-up, top-down, or hybrid development method without mathematical grounding. The majority of mainstream domain-independent and domain-specific multimedia ontologies introduced in the past decade, with or without MPEG-7 alignment, lack complex role inclusion axioms and DL-safe rules, and are limited to terminological and assertional knowledge. Consequently, most multimedia ontologies are actually controlled vocabularies, taxonomies, or knowledge bases only, rather than fully-featured ontologies, and are not suitable for advanced multimedia reasoning. To address the above issues, concepts, roles, individuals, and relationships of the professional video production and broadcasting domains have been formally modeled using $\mathcal{SROIQ}^{(\mathcal{D})}$, one of the most expressive decidable description logics, and then the axioms translated into OWL 2. The vocabulary of the new ontology has been aligned with standards in a new concept and role hierarchy. To further improve expressivity, $\mathcal{SROIQ}^{(\mathcal{D})}$ has been combined with DL-safe rules, without sacrificing expressivity yet ensuring decidability by restricting rules to known individuals. Ongoing work is in progress to extend this core ruleset further to reach an even higher level of reasoning power.